%% file: root.tex
\documentclass[letterpaper, 10 pt, conference]{ieeeconf}  

\IEEEoverridecommandlockouts                              

\usepackage{amsmath,amssymb,amsfonts}
\usepackage{graphicx}
\usepackage{textcomp}
\usepackage{xcolor}
\usepackage[most]{tcolorbox}
\usepackage{booktabs}
\usepackage{url}
\usepackage{hyperref}

\usepackage{colortbl}

\usepackage{caption}
\usepackage{subcaption}
\usepackage[utf8]{inputenc}

\newcommand \ignore[1]{}

\title{\LARGE \bf
Are Neuro-Inspired Multi-modal Vision-Language Models Resilient to Privacy Leakage?
}
\title{\LARGE \bf
Are Neuro-Inspired Multi-Modal Vision-Language Models Resilient to Membership Inference Privacy Leakage?
}

\author{
\begin{tabular}{c}
David Amebley$^{1}$ \quad Sayanton Dibbo$^{1,2}$\\[0.5ex]
\small $^{1}$The University of Alabama\\
\small {\tt dkamebley@crimson.ua.edu, sdibbo@ua.edu}\\[0.5ex]
\small $^{2}$Affiliate,
Alabama Center for the Advancement of AI and Director, Trustworthy AI Lab
\end{tabular}
\thanks{Department of Computer Science, The University of Alabama, 248 Kirkbride Ln., Tuscaloosa, AL 35401 USA.}
}

\begin{document}

\maketitle
\thispagestyle{empty}
\pagestyle{empty}

\begin{abstract}

In the age of agentic AI, the growing deployment of multi-modal models (MMs) has introduced new attack vectors that can leak sensitive training data in MMs, causing \textit{privacy leakage}. This paper investigates a \textit{black-box} privacy attack, i.e., \textit{membership inference attack (MIA)} on multi-modal vision-language models (VLMs). 
\textit{State-of-the-art} research analyzes privacy attacks primarily to unimodal AI/ML systems, while recent studies indicate MMs can also be vulnerable to privacy attacks. While researchers have demonstrated that biologically inspired neural network representations can improve unimodal models' resilience against adversarial attacks, it remains unexplored whether neuro-inspired MMs are resilient against privacy attacks.
In this work, we introduce a systematic neuroscience-inspired topological regularization (i.e., $\tau$-regularized) framework to analyze MM VLMs' resilience against image-text-based inference privacy attacks. We examine this phenomenon using three VLMs: BLIP, PaliGemma 2, and ViT-GPT2, across three benchmark datasets: COCO, CC3M, and NoCaps.
Our experiments compare the resilience of baseline and neuro VLMs (with topological regularization), where the $\tau > 0$ configuration defines the \textsc{neuro} variant of each VLM under varying values of the topological coefficient $\tau$ (0–3). We show how $\tau$-regularization affects the MIA attack success, offering a quantitative perspective on the privacy-utility trade-offs.
Our results on the BLIP model using the COCO dataset illustrate that MIA attack success in \textsc{neuro} VLMs drops by $\sim$ $24\%$ mean ROC-AUC, while achieving similar model utility (similarities between generated and reference captions) in terms of MPNet and ROUGE-2 metrics. This shows neuro VLMs are comparatively more resilient against privacy attacks, while not significantly compromising model utility.
Our extensive evaluation with PaliGemma 2 and ViT-GPT2 models, on two additional datasets: CC3M and NoCaps, further validates the consistency of the findings. This work contributes to the growing understanding of privacy risks in MMs and provides empirical evidence on neuro VLMs' privacy threat resilience.

\end{abstract}

\input{Sections/introduction}
\input{Sections/preli}

\input{Sections/method}

\input{Sections/experiments}

\input{Sections/results}
\input{Sections/conclusion}
\bibliographystyle{ieeetr}
\bibliography{ref}


\end{document}

%% file: Sections/introduction.tex
\section{Introduction}
\label{sec:intro}
In the age of Agentic AI, multi-modal models (MMs) developed with multiple data modalities and sources are becoming more popular in application areas, including autonomous robotics~\cite{cui2024survey,sreeram2025probing}, healthcare~\cite{yildirim2024multimodal, vengatachalam2025transforming}, education~\cite{doveh2024towards}, biometric authentication~\cite{vhaduri2024mwiotauth, muratyan2022opportunistic}, and content creation~\cite{behravan2024generative, cheng2025text}. These MMs often include different modalities, including image and text modalities.
\begin{figure*}[]

\begin{subfigure}{.5\textwidth}
  \centering
    \includegraphics[width=6.5cm, height=2.5cm]{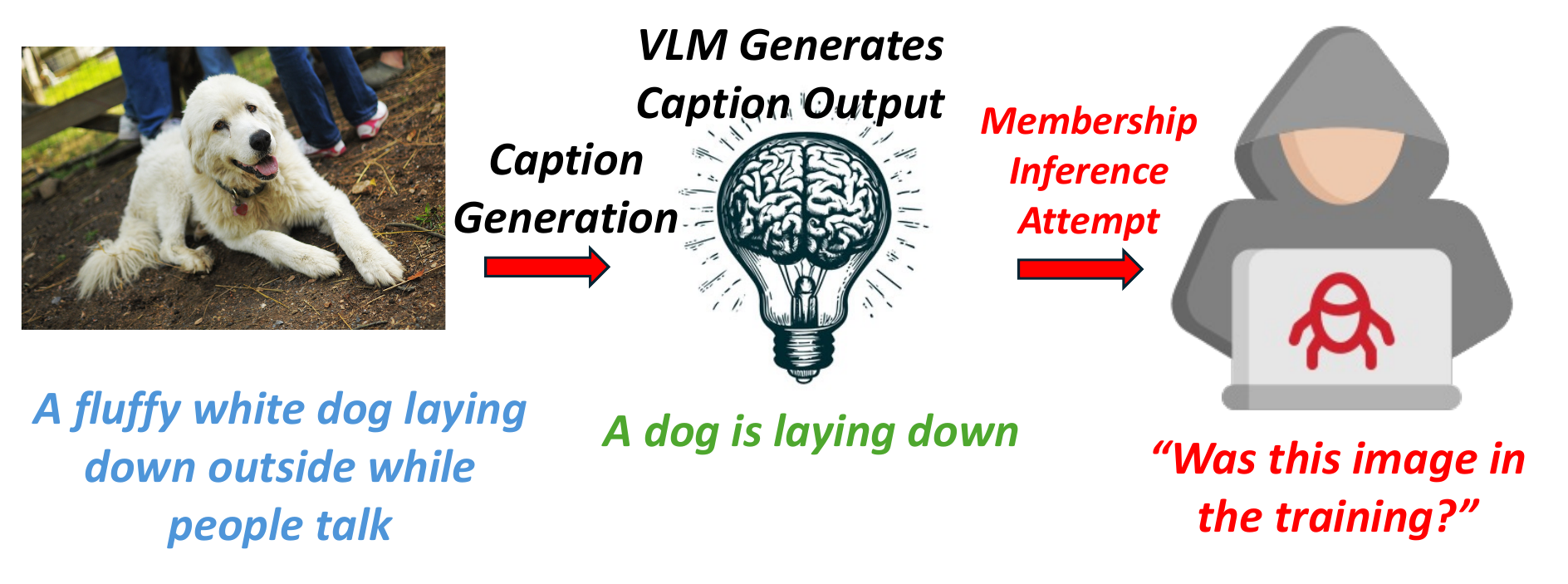}
  \caption{MIA Attack Overview}
        \label{fig:vlm_attack_overview}
\end{subfigure}
\hfill
\begin{subfigure}{.5\textwidth}
  \centering
    \includegraphics[width=7.4cm,  height=3.5cm]{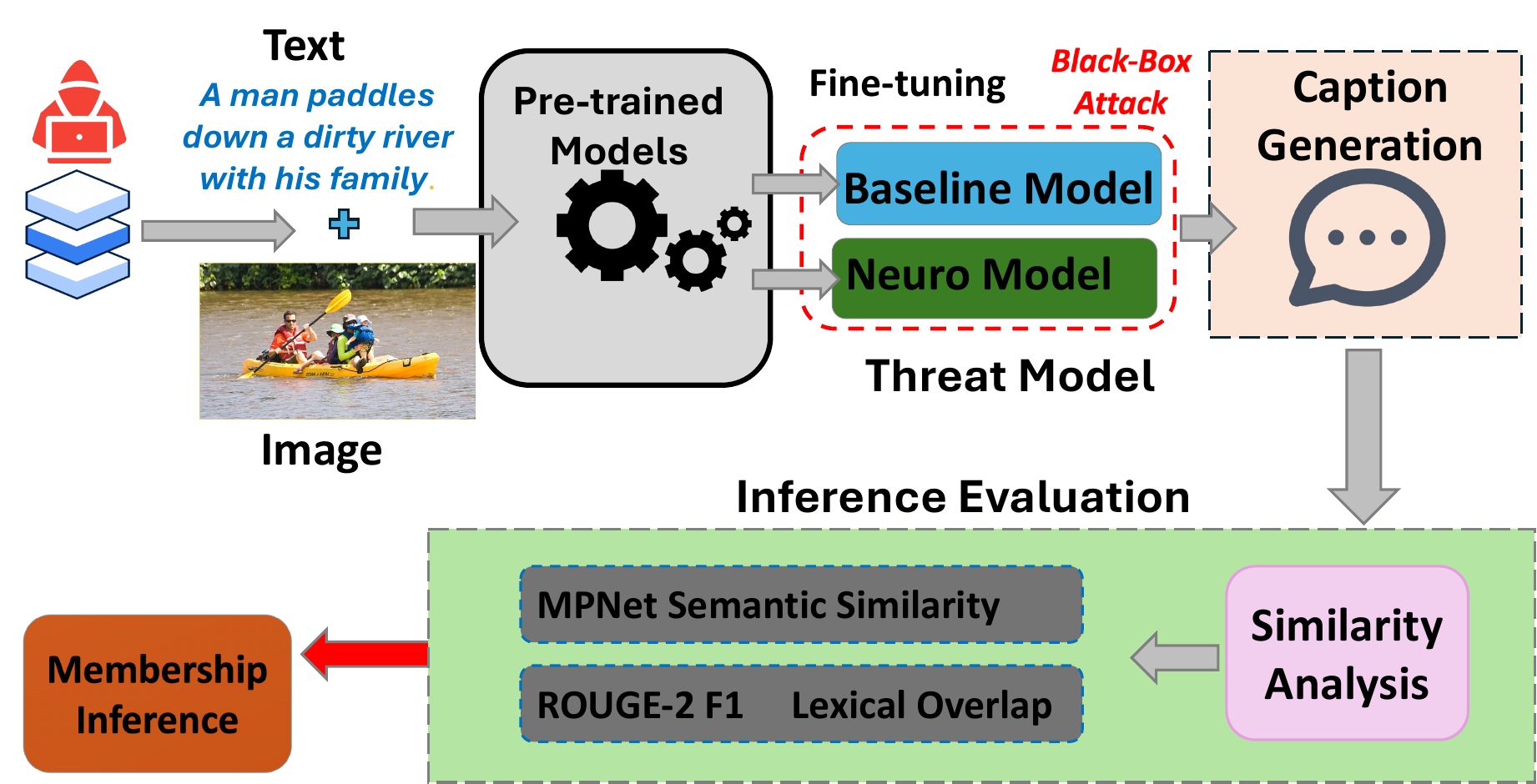}
        \caption{\textcolor{black}{MIA Pipeline on VLM Threat Model.}}
  \label{fig:vlm_overview_threat_model}
\end{subfigure}%
\caption{\textcolor{black}{In (a), we illustrate an overview of MIA attack on VLM, where an adversary infers whether a sample is part of training data by generating the model’s (VLMs) caption (output) from VLMs with queries. In (b), we present the MIA pipeline for neuro-inspired VLMs, in which the adversary fine-tunes pre-trained VLMs with topological regularization ($\tau$) and generates captions. Finally, the adversary decided the membership of a particular sample based on semantic (MPNet) and lexical (ROUGE-2) similarity measures between original and generated captions.}}
\label{fig:bothdiagrams}
\end{figure*}
   \textcolor{black}{In the last few years, Large Language Models (LLMs) like ChatGPT have seen more improvements and can now solve a wide range of problems, making them more popular than ever \cite{chang2024survey, bordes2024introduction}. These models are known for their language understanding, built on the foundation of Natural Language Processing (NLP) \cite{bordes2024introduction}. VLMs benefit from advances in LLM research and are considered part of the AI model cluster that leverages NLP and Computer Vision to create a more complete understanding of visual and textual information~\cite{li2024vision}. VLMs with this power can identify objects in an image and provide a natural language description of the object identified \cite{li2024vision, dai2023instructblip, lee2024vhelm}. VLMs are integrated into many applications and tools from basic to sophisticated tasks, e.g., extracting text from an image, explaining a diagram, or counting objects \cite{laurenccon2024matters, lee2024vhelm, li2024vision, liu2025survey}. This can further be incorporated in wearable computing~\cite{lien2024explaining}, trustworthy AI/ML computing~\cite{dibbo2024lcanets++}, health tracking~\cite{vhaduri2022predicting}, mission-critical agentic AI solutions~\cite{rolf2024mission}, and user authentication~\cite{vhaduri2023bag}.}

    \textcolor{black}{ Despite their capabilities, likewise unimodal models, VLMs are susceptible to vulnerabilities like adversarial and privacy attacks, especially due to their multi-modality~\cite{liu2025survey, laurenccon2024matters, zhao2023evaluating, ye2025survey, cheng2024unveiling, dibbo2023sok, ren2025seeing, li2025benchmark}. Among these is a type of attack in which image and text pairs are subtly altered to mislead the model to produce harmful outputs \cite{liu2025survey, ye2025survey, liang2025vl, ni2024physical, xu2024shadowcast}. Another common vulnerability in the form of an attack to VLMs is membership inference attacks (MIAs), where attackers (also known as adversaries) try to deduce whether the model used some specific data in its training, which can be a privacy risk~\cite{hu2025membership, choquette2021label, hu2023quantifying, li2020membership, li2021membership, shokri2017membership}.}
    
    \textcolor{black}{Shokri et al.~\cite{shokri2017membership} implemented and tested the first-ever \textit{black-box} membership inference attacks (MIAs) on machine learning models. \cite{hu2022membership, olatunji2021membership} affirmed the success of these attacks on different machine learning models, including classification and generative models, convolution neural networks (CNNs), and multilayer perceptron (MLP). Some of these attacks also target ML models that learned on multiple and sensitive image and text data~\cite{he2021node}.
    Recently, work on MIAs has expanded to LLM-based models, such as generative text LLMs. Some \textit{black-box}, text-only MIA methods in LLMs, such as the Repeat and Brainwash techniques, check for membership by examining patterns in the text the model produces \cite{wen2024membership, das2025security}. These methods are similar to MIAs on VLMs, like our work, in that they rely on measuring the semantic similarity or lexical overlap between generated captions and ground-truth data. However, the LLM attacks target text data rather than text-image pairs.
    More recently, we have seen some emerging work on MIAs on VLMs, with the first-ever by~\cite{hu2022m, li2024membership}, who explored MIAs on multi-modal image captioning systems using metric-based and feature-based similarity methods. Hu et al.~\cite{hu2025membership} highlighted the need for more attention to MIAs on VLMs, as most work on VLMs has focused on improving the performance of these models' interactions~\cite{deb2025toponets}. In their work, they proposed the first temperature-based MIA methods targeting instruction-tuning data in VLMs.}

    \textcolor{black}{Although researchers started focusing on privacy vulnerabilities to VLMs, it has generally focused on detecting MIAs on regular VLMs using metric- or similarity-based signals~\cite{wu2022membership, li2024membership}, which exploit the proximity of the model-generated caption to the ground-truth training caption. These efforts consider regular VLMs vulnerabilities, which do not seem to be resilient against MIAs. Studies, on the other hand, pointed out that neuroscience-inspired biologically regularized unimodal models can be more resilient to privacy threats compared to regular neural networks due to their capabilities to capture subtle changes in samples~\cite{dibbo2024improving, teti2022lcanets}. However, it is not yet explored how this holds for multi-modal complex VLMs. Therefore, we focus on developing a biologically inspired regularization-based framework to assess the MIA vulnerabilities on MM VLMs.    
    We explore how neuroscience-inspired model-level structural regularization, specifically $\tau$-regularization (topology-inducing training)~\cite{deb2025toponets}, can impact VLMs resilience. Instead of just detecting data leakage through output metrics, we study an internal framework that makes models more resilient against MIAs, without compromising performance (utility).}  
    \textcolor{black}{More importantly, to the best of our knowledge, no research has examined how neuro-inspired vision-language models (i.e., those trained with topographic regularization) affect privacy. This situation leaves an important gap in understanding whether topographic regularization can affect memorization and membership inference attacks.}
    
Our main contributions in this work are as follows: \par
\begin{itemize}
    \item We introduce the topological regularization framework (i.e., $\tau$-regularized) to fine-tune MM VLMs (i.e., neuro-inspired models)
    \item We study resilience of neuro-inspired VLMs for the first time on \textit{black-box} \textit{membership inference} privacy attacks
    \item We extensively evaluate the neuro-inspired VLMs' privacy leakage in three different datasets (COCO, CC3M, and NoCaps) with three different VLMs (BLIP, PaliGemma 2, and ViT-GPT2) and compare the neuro-inspired VLMs' resilience
    \item We present quantitative comparisons between the privacy attack success rates (accuracy) and VLMs' performance (caption similarity) among neuro-inspired VLMs and baseline models, illustrating the privacy vs. performance trade-offs
    \item We ablate the neuro-inspired VLMs with different $\tau$ values to capture pattern of VLMs resilience against privacy attacks 
\end{itemize}
\textbf{Organization:} We organize the remainder of this paper as follows:
Section~\ref{sec:preli} explains Membership Inference Attacks and Vision-Language Models. Section~\ref{sec:methods} describes our methodology, including the \textit{black-box} threat model and the $\tau$-regularized fine-tuning process. Section~\ref{sec:exp} outlines the experimental setup, the models used, and the evaluation metrics. Section~\ref{sec:results} discusses the results and the resilience of neuro-inspired VLMs. Finally, Section~\ref{sec:conclusion} concludes the paper and offers ideas for future research.

%% file: Sections/preli.tex
\section{Preliminaries}
\label{sec:preli}
In this section, we illustrate preliminaries on the multi-modal vision-language models (VLMs) and membership inference attacks (MIAs).

\subsection{\textcolor{black}{Membership Inference Attacks (MIAs)}}\label{subsec:mia}
Before we define inference attacks, we first want to examine privacy in the context of vision language and machine learning models. 
Dalenius et al. claimed that a model should not show more about its input than what we would have known about the input without the model \cite{dalenius1977towards}. 
\par
An adversary can persistently query a model to determine whether certain data was used to train it or not. The attacker usually has some dataset that they \emph{suspect} may have been used to train the model \cite{carlini2022membership,rezaei2021difficulty,hayes2017logan,liu2022membership,song2019membership,zhang2021membership,li2021membership}. During this time, they feed the model with the data they have and observe the model's output \cite{hu2022membership,hu2023defenses,truex2018towards}. 
Formally, we can express this as:

\begin{equation}\label{eqn:mia}
\mathcal{A} : (u, v) \mapsto 
\begin{cases}
0, & \text{if } (u, v) \notin \mathcal{D},\\[4pt]
1, & \text{if } (u, v) \in \mathcal{D},
\end{cases}
\end{equation}

\noindent where \( u \) is an input instance, \( v \) is its associated label or caption, $\mathcal{D}$ represents the model's training dataset, and $\mathcal{A}$ outputs \( 0 \) or \( 1 \) to indicate whether \( u,v \) is inferred to be a non-member or member of the training set, respectively~\cite{liu2025efficient}.

A strong positive response infers membership, i.e., the model has likely seen the data before. If an attacker knows that some data was used to train a model, it can result in a privacy leak \cite{shokri2017membership}. 
Various membership inference methods have been tested in the past, including shadow model-based MIAs~\cite{shokri2017membership, hu2022membership, truex2018towards,gong2022private, liu2022ml}. 

\textit{White-box} membership inference attacks exploit internal signals, such as gradients, activations, or model parameters, to detect overfitting on specific data samples. They are usually more effective than \textit{black-box} attacks because they leverage the model's training dynamics \cite{hu2022membership, hu2022membership, hu2022membership, pang2023white, wu2023understanding, leino2020stolen, sablayrolles2019white}. 
\textit{Black-box} membership inference attacks use model outputs to infer membership. Unlike \textit{white-box} attacks, these types have no access to the model's internal details, such as activations or gradients. They are widely used in ML-as-a-Service scenarios where users lack internal access to the model \cite{shokri2017membership, hayes2017logan,hu2022membership}. 

\subsection{\textcolor{black}{Vision-Language Models (VLMs)}}\label{subsec:vlms}
Vision-language models (VLMs) are models that learn to connect images and text by using large amounts of image and text data. They can understand natural language and reason on different types of information.

\begin{equation}\label{eqn:vlm}
\mathcal{L}_{\mathrm{VLM}}
=
- \log
\frac{
\exp\!\left( \mathrm{sim}(z_i^{\mathrm{img}},\, z_i^{\mathrm{txt}}) / \beta \right)
}{
\sum_{j} \exp\!\left( \mathrm{sim}(z_i^{\mathrm{img}},\, z_j^{\mathrm{txt}}) / \beta \right)
},
\end{equation}

\noindent
where $z_i^{\mathrm{img}}$ is the embedding of the $i$-th image from the vision encoder,
and $z_i^{\mathrm{txt}}$ is the embedding of its paired caption from the language encoder.
The function $\mathrm{sim}(\cdot,\cdot)$ measures how similar two embeddings are
(e.g., cosine similarity), and $\beta$ is a temperature constant.
This loss pushes each image embedding to be most similar to its matching caption
compared to other captions in the batch~\cite{zhang2024vision, li2020oscar}. 

Regarding their strengths, VLMs connect visual information with text to make it easier to understand images and their descriptions \cite{zhang2024vision}. Because VLMs learn from large-scale image–text pairs, they generalize with little or no new training \cite{zhang2024vision, shinde2025survey}. They also perform classification, captioning, visual question answering (VQA), text recognition (OCR), and information retrieval following instructions \cite{lee2024vhelm}. Methods like freezing encoders, compressing tokens, and adding adapters can make VLMs more efficient while retaining their strong performance \cite{shinde2025survey}.


%% file: Sections/method.tex
\section{Methodology}
\label{sec:methods}

We consider a \textit{black-box} MIA scenario, where the adversary attempts to determine whether a specific image-text pair was part of the training dataset of a deployed VLM.

\subsection{\textcolor{black}{Adversarial Capabilities}}\label{subsubsec:advers_capab}
Adversarial capabilities determine the effectiveness of a membership inference attack and influence the attacker's chances of success. Previous research classifies adversaries based on the amount of information they can access from the target model \cite{wu2024rethinking, chobola2023membership, hu2025membership, wu2021adapting}. 
In \emph{white-box} settings, attackers can access internal model parameters, gradients, or their hidden representations. In \emph{soft black-box} settings, attackers can query the model and receive intermediate outputs, such as logits or probability distributions. The most restrictive and realistic scenario for deployed VLMs is the \emph{strict black-box} setting. With this scenario, adversaries can only see the final output of the model \cite{li2020practical, woitschek2021physical, ter2024adversarial}. 
In our threat model, the adversary operates under this strict \textit{black-box} assumption. They can submit an input image and only see the generated caption, with no access to model weights, training data, embeddings, or confidence scores. This scenario is common with deployed VLMs with only inference APIs access~\cite{truex2019demystifying, hu2022membership, zhang2021membership, liang2024model, tramer2016stealing, gong2021model, liu2021machine}.

\subsection{Attack Overview} \label{subsec:attack_overview}
The adversary's goal is to find out if a specific image-caption pair \((u,v)\) was used to train the target VLM \(\mathcal{M}_\tau\). In this \textit{black-box} threat model, the adversary can only send an image \(u\) to \(\mathcal{M}_\tau\) and see the caption it produces, which we call \(v' = \mathcal{M}_\tau(u)\), as described in Section~\ref{subsubsec:advers_capab}.

In the setup shown in Fig.~\ref{fig:vlm_attack_overview}, the attacker queries the model with an image and receives a caption. They then compare this caption to the image's ground-truth caption. If the caption is very similar to the original text, the adversary infers that \((u,v)\) likely belongs to the training. If it is not, they classify it as a non-member.
To measure caption similarity, they can use two methods: (i) MPNet embedding similarity, which checks for semantic alignment, and (ii) ROUGE-2 overlap for lexical matching. These two methods combine to create the membership score.

\subsection{Attack Pipeline on $\tau$-Regularized VLMs}
\label{subsec:attack_pipeline}
Fig.~\ref{fig:vlm_overview_threat_model} shows the detailed steps of the attack used in our evaluation. Although the attacker does not modify the model or train it, we analyze how different levels of topographic regularization, $\tau \in \{0,2,3\}$, can impact privacy. For each level of $\tau$, the attack follows these steps: 

(i) \textbf{Query.} The attacker sends an input image \(u\) to the model $\mathcal{M}_\tau$ and gets a generated caption \(v'\).
\par
(ii) \textbf{Similarity Computation.} The attacker checks how similar \(v'\) is to the ground-truth caption(s) using MPNet semantic similarity and ROUGE-2 lexical overlap. This gives us two indicators of membership:
 \[ s_{\mathrm{mp}}(u) = \mathrm{sim}_{\mathrm{MPNet}}(v', v), \quad s_{\mathrm{rg}}(u) = \mathrm{sim}_{\mathrm{ROUGE2}}(v', v). \]

We describe similarity metrics in Section~\ref{subsec:eval_metrics}.

\par
(iii) \textbf{Membership Decision.} If the similarity is high, it suggests the model memorized training information, leading the attacker to infer membership. On the other hand, a lower score will indicate likely non-membership.
These steps connect directly to Fig.~\ref{fig:vlm_overview_threat_model}, which illustrates how captions are generated from $\mathcal{M}_\tau$, compared to references, and then converted to membership predictions using similarity scores, which explain how the attacker interacts with the model.

\subsection{Adversarial Goal} \label{subsubsec:advers_goal}
We formalize this \textit{black-box} attack capability as follows. 

\begin{equation}\label{eqn:advers_capb_vlm}
\mathcal{M}_{\tau} : \mathcal{U}\rightarrow \mathcal{V}
\end{equation}

represent a VLM fine-tuned under topological regularization coefficient $\tau \ge 0$, where $\mathcal{U}$ is the \emph{image domain} and $\mathcal{V}$ is \emph{the caption domain}. The adversary can submit an image $u\in\mathcal{U}$ to $\mathcal{M}_{\tau}$ and observe its generated caption

\begin{equation}\label{eqn:advers_capb_captn}
v' = \mathcal{M}_\tau(u).
\end{equation}

\noindent\textit{Here, $\mathcal{M}_\tau$ refers to the deployed VLM, $u$ is the query image, and $v'$ is the caption that the model generates. The adversary cannot access any gradients, embeddings, or model parameters.}
For a given query image \( u \), the attacker attempts to infer whether \( u \) was part of the model's private training set $\mathcal{D}_{\text{train}}$ \cite{shokri2017membership, ko2023practical}.
Formally, the goal is to infer the membership indicator \cite{jegorova2022survey}
\begin{equation}\label{eqn:advers_goal}
b(u) = \begin{cases} 1 & \text{if } u \in \mathcal{D}_{\text{train}},\\ 0 & \text{otherwise} \end{cases}
\end{equation}

To clarify the membership inference problem, we specify the training objective of the model being attacked. When fine-tuning, we optimize the VLM using a topography-regularized loss:
\begin{equation}\label{eqn:topoloss}
\mathcal{J}_{\tau} = \mathcal{J}_{\text{cap}} + \tau\,\mathcal{R}_{\text{topo}},
\end{equation}
where $\mathcal{J}_{\text{cap}}$ is the standard captioning loss (cross-entropy), and $\mathcal{R}_{\text{topo}}$ is the topographic regularizer introduced by Deb et al.~\cite{deb2025toponets}. The regularizer encourages smoother, spatially organized internal representations by aligning each feature map with a blurred counterpart. Formally, for cortical maps $C$ and their blurred versions $C'$, 
\begin{equation}
\mathcal{R}_{\text{topo}} 
= -\frac{1}{N} \sum_{i=1}^{N}
\frac{C_i \cdot C'_i}{\lVert C_i \rVert\, \lVert C'_i \rVert}.
\end{equation}

Unlike cross-entropy, which only focuses on improving task performance, $\mathcal{R}_{\text{topo}}$ adds a structural rule, i.e, inductive bias, that shapes how the model understands information. Parameter $\tau$ controls inductive bias strength~\cite{deb2025toponets}.

\subsection{Information Available to the Adversary} \label{subsubsec:info_avail_advers}
We assume that the adversary has a set of candidate sample images, which they may have drawn from a distribution similar to the training data. For each query image \( u \), the adversary also assumes access to an accompanying reference caption \( v \), e.g., publicly available annotations \cite{lin2014microsoft, srinivasan2021wit, schuhmann2022laion}. These candidate samples are used to query the model, compute similarity statistics between \( v \) and $v' = \mathcal{M}_\tau(u)$, and derive a membership decision rule $\hat{b}(u)$.

\subsection{Adversarial Knowledge Assumptions} \label{subsubsec:advers_knowledge_assump}
In this attack, we assume that the adversary knows the targeted model is a VLM accessible via query \cite{carlini2022membership, correia2018copycat, tramer2016stealing}. The adversary also knows public data distributions resembling the training domain exist \cite{sharma2018conceptual, lin2014microsoft, agrawal2019nocaps, radford2021learning, schuhmann2022laion}. Additionally, we assume that the adversary is aware of the availability of a similarity scoring function \( S \) \cite{reimers2019sentence}.
Importantly, we also infer that the adversary is not exposed to the internal parameters or gradients of the model \cite{shokri2017membership, yeom2018privacy}.
Again, we consider that they do not know the \emph{exact} samples used in training \cite{carlini2022membership}.
We also assume that the adversary is unaware of the optimizer state or hyperparameters used. This scenario is common with cloud-API threats \cite{carlini2022membership, tramer2016stealing}.

\subsection{Threat Model} \label{subsec:t_model}
The adversary assesses how closely the generated caption \( v' \) matches the reference caption \( v \) using similarity metrics like MPNet \cite{pavlyshenko2025semantic, song2020mpnet} and ROUGE-2 \cite{barbella2022rouge, lin-2004-rouge}. We denote a similarity function as \cite{carlini2023extracting, witschard2022interactive}:
\begin{equation}
S : \mathcal{V} \times \mathcal{V} \to \mathbb{R}.
\end{equation}

For a \emph{query image} \( u \):
\begin{equation}
s(u) = S\big(v,\, \mathcal{M}_\tau(u)\big),
\end{equation}

The adversary performs a threshold-based attack
\begin{equation}
\hat{b}(u)= \begin{cases} 1 & \text{if } s(u) \ge t,\\ 0 & \text{otherwise}, \end{cases}
\end{equation}
where \( t \) is chosen to maximize discrimination between members and nonmembers.
Here, \(u\) denotes an input image, and \(v\) is its reference caption. \(v' = \mathcal{M}_\tau(u)\) represents the generated caption, whereas 
\(S(\cdot,\cdot)\) computes semantic similarity between the generated and reference captions. 
\(s(u)\) is the resulting similarity score; 
and \(b(u) \in \{0,1\}\) indicates whether \(u\) is a training member.
For each model-dataset configuration, we compute
\begin{equation}
\alpha_{\text{in}},~\alpha_{\text{out}}
\end{equation}
The mean similarity scores for the member and nonmember samples, therefore, become as follows:
\begin{equation}\label{eqn:mean_sim_mem_nonmem}
\alpha_{\text{in}} = \mathbb{E}_{u\in \mathcal{D}_{\text{train}}}[s(u)], \qquad \alpha_{\text{out}} = \mathbb{E}_{u\notin \mathcal{D}_{\text{train}}}[s(u)]
\end{equation}
We denote the similarity gap as follows:
\begin{equation}\label{eqn:sim_gap}
\Delta_\tau = \alpha_{\text{in}} - \alpha_{\text{out}}.
\end{equation}
Larger $\Delta_\tau$ implies a stronger membership signal \cite{salem2018ml, yeom2018privacy, shokri2017membership}. Under $\tau$-regularization, we expect the $\Delta_\tau$ to shrink.
Membership inference success is computed by comparing $\hat{b}(u)$ to ground-truth \( b(u) \) \cite{hayes2017logan}.

\subsection{Topographic ($\tau$-Regularization) Model} \label{subsubsec:topo_regul}
To improve membership privacy, we consider VLMs trained under topological regularization~\cite{deb2025toponets}.

\begin{equation}\label{eqn:topo_regularization}
\mathcal{J}_{\tau} = \mathcal{J}_{\text{cap}} + \tau \cdot \mathcal{R}_{\text{topo}}.
\end{equation}

where \(\mathcal{J}_{\text{cap}}\) is the standard captioning loss,
and \(\mathcal{R}_{\text{topo}}\) encourages nearby features in the model to learn related functions.
The scalar \(\tau\) determines how strongly this topological constraint is applied~\cite{maxwell2020mapping, chitra2025mapping, hammer2010topographic} (explicit form of \(\mathcal{R}_{\text{topo}}\) is presented in Section~\ref{subsubsec:advers_goal}).

As \(\tau\) increases, the learned representations become more organized ("Neuro") \cite{ handjaras2015topographical, patel2014topographic, zhu2018ldmnet, santos2022avoiding, nusrat2018comparison, moradi2020survey, wu2017reducing}, which should reduce the model's reliance on memorizing specific training examples and therefore mitigate membership leakage \cite{deb2025toponets, tivno2006dynamics, hyvarinen2001topographic, hu2023defenses, kaya2020effectiveness, hu2021ear, hu2022membership}.
In this work, we study three configurations of the model: 
the baseline version (\(\tau = 0\)), 
a Neuro-regularized version (\(\tau = 2\)), 
and a stronger Neuro++ configuration (\(\tau = 3\)). 
These settings allow us to examine how increasing \(\tau\) affects privacy leakage.

\begin{table*}[t]
\vspace{4em}
\centering
\caption{
\textcolor{black}{Performance comparison among \textsc{Baseline} and $\tau$-regularized neuroscience-inspired models (\textsc{Neuro} with $\tau=2$ and \textsc{Neuro++} with $\tau=3$) on the COCO dataset. The table reports member and non-member similarity scores for MPNet and ROUGE-2, along with the resulting ROC-AUC values (mean$\pm$std) across BLIP, PaliGemma 2, and ViT-GPT2.
}}
\begin{tabular}{ |p{1.0cm}| p{1.8cm}| p{1.8cm}| p{1.8cm}| p{1.8cm}|  p{1.8cm}| p{1.8cm}| p{2.0cm}|  }
\hline
 Dataset  &Model  & Threat Model   & MPNet  $\downdownarrows$ & MPNet  $\downdownarrows$ &ROUGE-2 $\downdownarrows$ &ROUGE-2$\downdownarrows$& ROC-AUC$\upuparrows$\\
   &  &    & (Member)   & (Non-Member)  &(Member) &(Non-Member) & \\
\hline

COCO &BLIP  &\textsc{Baseline}  &$0.723$ &$0.663$ & $0.249$ & $0.171$ &$94.00 \pm 9.90$\\ 
 & &\textbf{\textsc{Neuro}} & $\mathbf{0.797}$ & $\mathbf{0.793}$	&$\mathbf{0.425}$	 &$\mathbf{0.380}$ & $\mathbf{70.88 \pm 18.95}$ \\ 
\rowcolor{gray!50}
 & &  \textbf{\textsc{Neuro++}} & $\mathbf{0.698}$ & $\mathbf{0.687}$	&$\mathbf{0.319}$ &$\mathbf{0.312}$	 & $\mathbf{63.46 \pm 10.88}$ \\ 
\hline

\hline
\hline 
 &PaliGemma 2  &\textsc{Baseline}  &$0.605$ &$0.603$ &$0.227$ &$0.203$ &$70.86 \pm 19.53$\\ 
 & &\textbf{\textsc{Neuro}} & $\mathbf{0.717}$ & $\mathbf{0.712}$	&$\mathbf{0.111}$ &$\mathbf{0.102}$	 & $\mathbf{69.98 \pm 13.48}$ \\ 
\rowcolor{gray!50}
 & &  \textbf{\textsc{Neuro++}} & $\mathbf{0.735}$  & $\mathbf{0.732}$ 	&$\mathbf{0.338}$   &$\mathbf{0.318}$	 & $\mathbf{66.76 \pm 13.57}$ \\ 
\hline

\hline
\hline 
 &ViT-GPT2  &\textsc{Baseline}  &$0.771$ &$0.773$ &$0.318$ &$0.304$ &$55.51 \pm 16.47$\\ 
 & &\textbf{\textsc{Neuro}} & $\mathbf{0.773}$  & $\mathbf{0.782}$	&$\mathbf{0.357}$    &$\mathbf{0.373}$	 & $\mathbf{29.38 \pm 10.58}$ \\ 
\rowcolor{gray!50}
 & &  \textbf{\textsc{Neuro++}} & $\mathbf{0.775}$  & $\mathbf{0.776}$	&$\mathbf{0.357}$    &$\mathbf{0.368}$	 & $\mathbf{38.39 \pm 10.28}$ \\ 
\hline
\end{tabular}
\label{table:performance_coco}
\end{table*}

%% file: Sections/experiments.tex
\section{Experiments}
\label{sec:exp}
In this section, we describe how we carry out our membership inference attacks on the pretrained VLMs. We outline the evaluation setup, models, datasets, caption-similarity pipeline, and attack formulation.
\emph{Our implementation is based on the following codebase.\footnote{The code can be found at \url{https://github.com/YukeHu/vlm\_mia}.}}

\subsection{Experimental Setup}\label{subsec:exp_setup}
\paragraph{Access Model and Infrastructure.}
We tested pre-trained VLMs BLIP, ViT-GPT2, and PaliGemma 2 using locally saved Hugging Face checkpoints on the UA HPC cluster, which has an A100 GPU with 80 GB. The attacker can only input image-caption pairs and receive captions in return. They cannot access any internal parameters or gradients. In addition, there are no limits on the number of queries in our experiments.

\paragraph{Fine-Tuning.}
We fine-tuned each model for each dataset under three different topological regularization settings: \(\tau \in \{0, 2, 3\}\), which we call \emph{Baseline}, \emph{Neuro}, and \emph{Neuro++}. During fine-tuning, we update the model parameters while applying a \(\tau\)-scaled topographic penalty to the decoder features. We save a separate checkpoint directory for every combination of model, dataset, and \(\tau\).

\paragraph{Datasets and Splits.}
We use an 80/20 member split of each dataset for fine-tuning. We keep the non-member split for evaluation only. Full details about sampling appear in Section~\ref{subsec:datasets}.
\begin{figure*}[]
    \centering
    \begin{subfigure}[t]{0.46\linewidth}
        \centering
        \includegraphics[width=\linewidth]{Figures/images/food.jpg}
        \vspace{3pt}
        \small Caption ($\tau=0$): ``a close up of a tray of food with broccoli''
        \\
        \caption{Member, $\tau=0$}
    \end{subfigure}
    \hfill
    \begin{subfigure}[t]{0.46\linewidth}
        \centering
        \includegraphics[width=\linewidth]{Figures/images/food.jpg}
        \vspace{3pt}
        \small Caption ($\tau=3$): ``a group of containers filled with food sitting on top of a table''
        \\
        \caption{Member, $\tau=3$}
    \end{subfigure}
    \small{Reference caption: \textbf{``Closeup of bins of food that include broccoli and bread.''}}

    \vspace{8pt}

    \begin{subfigure}[t]{0.46\linewidth}
        \centering
        \includegraphics[width=\linewidth]{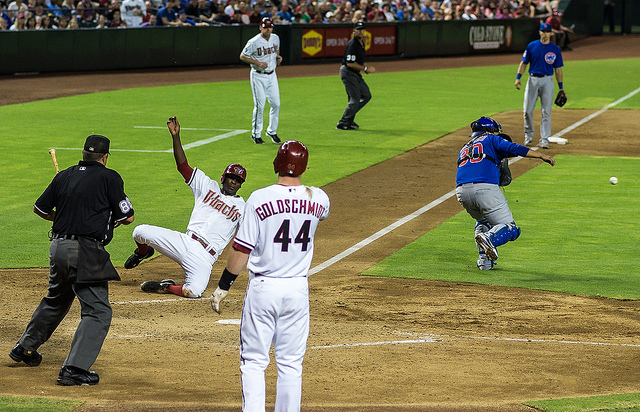}
        \vspace{3pt}
        \small Caption ($\tau=0$): ``a group of baseball players playing a game of baseball''
        \\
        \caption{Non-member, $\tau=0$}
    \end{subfigure}
    \hfill
    \begin{subfigure}[t]{0.46\linewidth}
        \centering
        \includegraphics[width=\linewidth]{Figures/images/baseball.jpg}
        \vspace{3pt}
        \small Caption ($\tau=3$): ``a baseball player is throwing a ball on a field''
        \\
        \caption{Non-member, $\tau=3$}
    \end{subfigure}
    \vspace{5pt}
    \small Reference caption: \textbf{``A baseball field full of baseball players standing on a field.''}

    \caption{
      Example image-caption pairs from COCO (BLIP) showing how captions change 
      under baseline ($\tau=0$) and neuro-inspired (topographically regularized, i.e., $\tau=3$) variant for member and non-member samples.
    }
    \label{fig:coco_blip_image_cap_pairs}
\end{figure*}

\paragraph{Caption Generation.}
We generate one caption per image. The ViT-GPT2 model uses greedy decoding (no sampling,  single beam, set to a \texttt{max\_new\_tokens=20}). The BLIP model uses its default \texttt{generate()} configuration, which also includes no sampling and a single beam. PaliGemma 2 generates captions with a brief \texttt{"Describe the image"} text prompt. We used nucleus sampling (\texttt{top\_p=0.9}), \texttt{temperature=0.7}, \texttt{repetition\_penalty=1.2}, and \texttt{max\_new\_tokens=60}. We remove any echoed prompt prefix before scoring.

\paragraph{Per-Configuration Evaluation.}
We calculate all similarity statistics (MPNet, ROUGE-2), similarity gaps, and membership decisions separately for each combination of model, dataset, and $\tau$ configuration. We use the member and non-member sets described earlier.

\subsection{Models Evaluated}\label{subsec:models}
We evaluate three representative vision-language models (VLMs): \textbf{BLIP} \cite{li2022blip}, \textbf{ViT--GPT2} \cite{radford2019language, dosovitskiy2020image}, and \textbf{PaliGemma\,2} \cite{beyer2024paligemma}. We also examine each model under three levels of topographic regularization:
\(\tau \in \{0, 2, 3\}\). The configuration \(\tau=0\) corresponds to the standard pretrained model, whereas \(\tau>0\) adds topographic regularization during supervised fine-tuning.
We fine-tune all models end-to-end using image-caption pairs. For each combination of model, dataset, and \(\tau\) configuration, we save a separate checkpoint. We use these checkpoints later to generate captions and check membership based on similarities. This approach enables us to compare the extent of privacy leakage across different models, datasets, and levels of regularization.

\subsection{Datasets}\label{subsec:datasets}
We evaluate our models on three image-caption datasets: \textbf{COCO}~\cite{lin2014microsoft}, \textbf{NoCaps}~\cite{agrawal2019nocaps}, and \textbf{CC3M}~\cite{sharma2018conceptual}. For each dataset, we randomly select 400 image-caption pairs and group them as \emph{members}, and an additional 400 pairs as \emph{non-members}, keeping a balanced 1:1 ratio. We ensure that these two sets are completely disjoint from each other. We fine-tune the models only on the member pairs, splitting into ${80\%}$ for training and ${20\%}$ for validation. The non-member pairs remain unseen during training, and we use them for evaluation. We apply the same sampling for all datasets.

\subsection{Similarity Scoring}\label{subsec:sim_scoring}
For each model, dataset, and $\tau$ configuration, we evaluate how closely a generated caption matches its reference. We calculate two scores: ROUGE-2 F1 \cite{barbella2022rouge, lin-2004-rouge} for lexical overlap and MPNet cosine similarity \cite{song2020mpnet} for semantic similarity. We use the MPNet model from sentence-transformers to generate embeddings with \texttt{normalize\_embeddings} set to \texttt{True}. For each image, we compare the generated caption with the reference captions and take the maximum score from the reference captions. For each configuration and membership category (member/non-member), we log the per-image similarity values, as well as the mean similarity and sample count.

\subsection{MIA Pipeline}\label{subsec:mia_pipeline}
We evaluate membership inference for each combination of dataset, model, regularization coefficient (\(\tau\)), and granularity parameter \(g\) (\(g \in \{10, 50, 100, 150, 200\}\)). Doing that helps us to observe how the attack success rate changes with different sampling depths. For each run, we load the generated captions for both member and non-member sets, with their matching reference captions. 
For each image, we calculate a membership signal that represents the maximum similarity between the generated caption and its available reference captions, using a single metric (MPNet or ROUGE-2) per run and producing a single score per image.
We do not use any shadow models in our setup. The attack compares the distributions of similarity scores from member and non-member sets and adjusts the decision thresholds accordingly. This method computes the ROC-AUC for these two score distributions, with a higher AUC indicating better attack success. We explain the ROC-AUC metric in Section~\ref{subsec:eval_metrics}.
\begin{figure*}
     \centering
     \begin{subfigure}[]{0.3\textwidth}
         \centering
         \includegraphics[width=\textwidth]{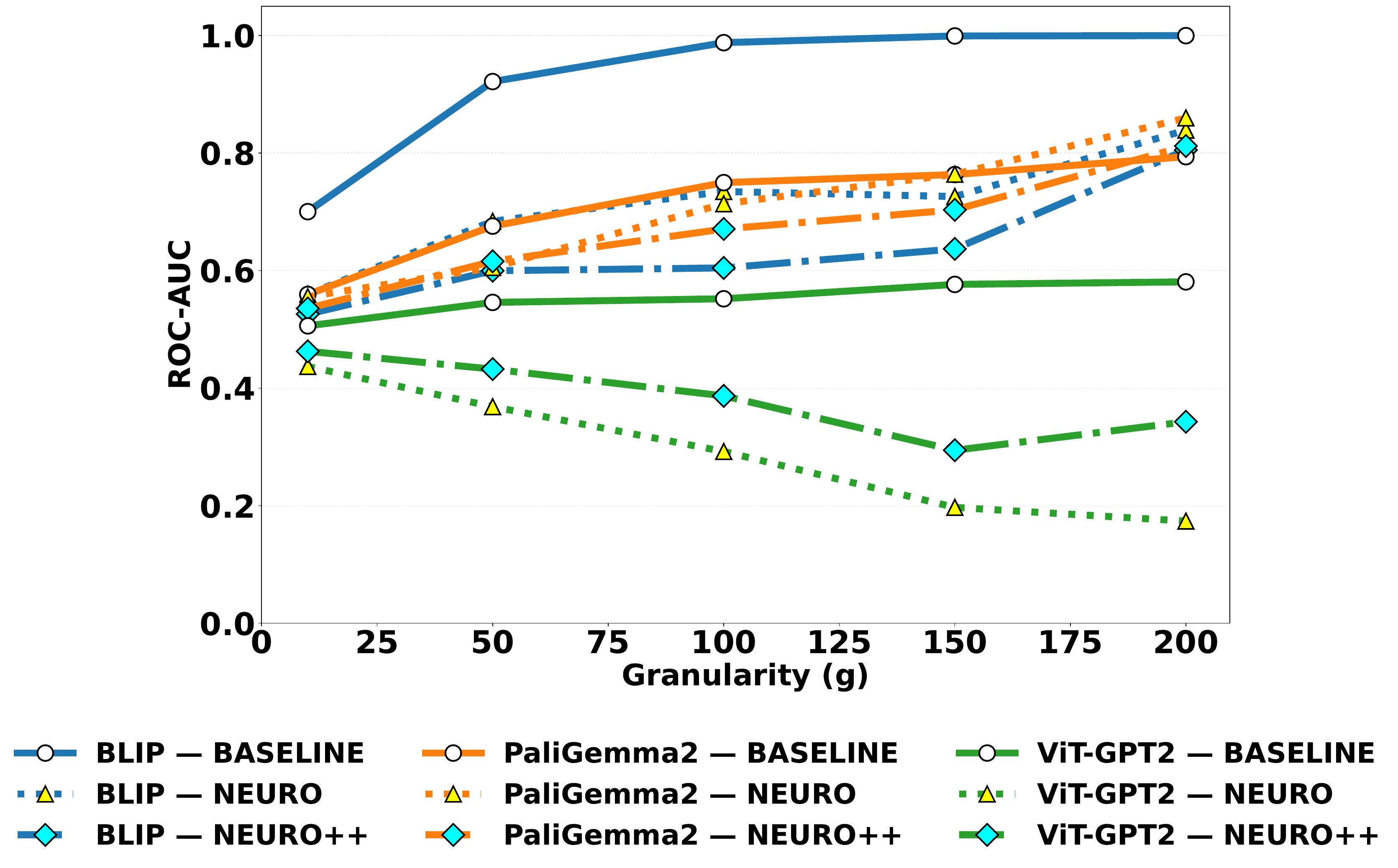}
         \caption{ROC-AUC with COCO Dataset}
         \label{fig:att_acc_coco}
     \end{subfigure}
     \hfill
     \begin{subfigure}[]{0.3\textwidth}
         \centering
         \includegraphics[width=\textwidth]{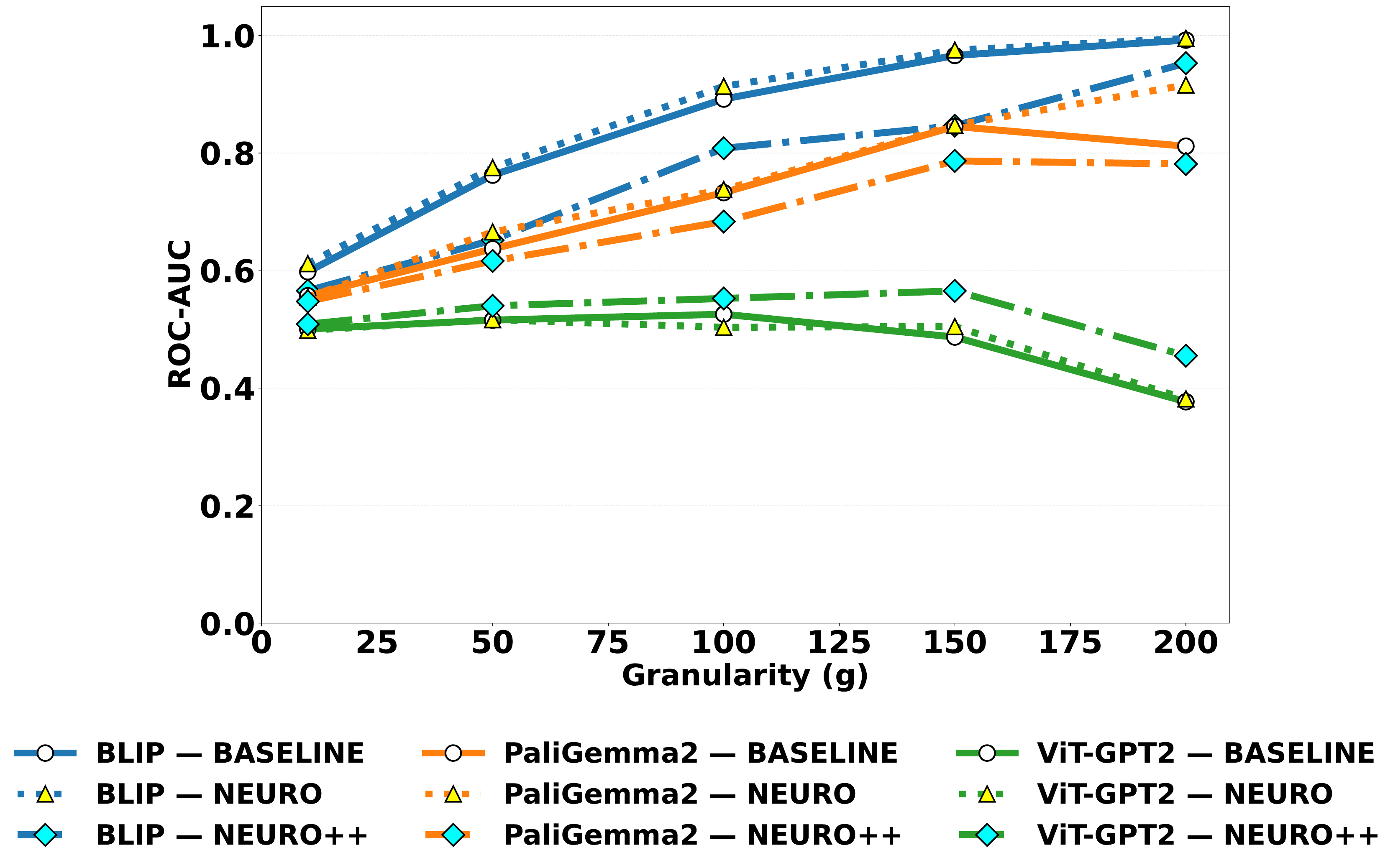}
         \caption{ROC-AUC with CC3M Dataset}
         \label{fig:att_acc_cc3m}
     \end{subfigure}
     \hfill
     \begin{subfigure}[]{0.3\textwidth}
         \centering
         \includegraphics[width=\textwidth]{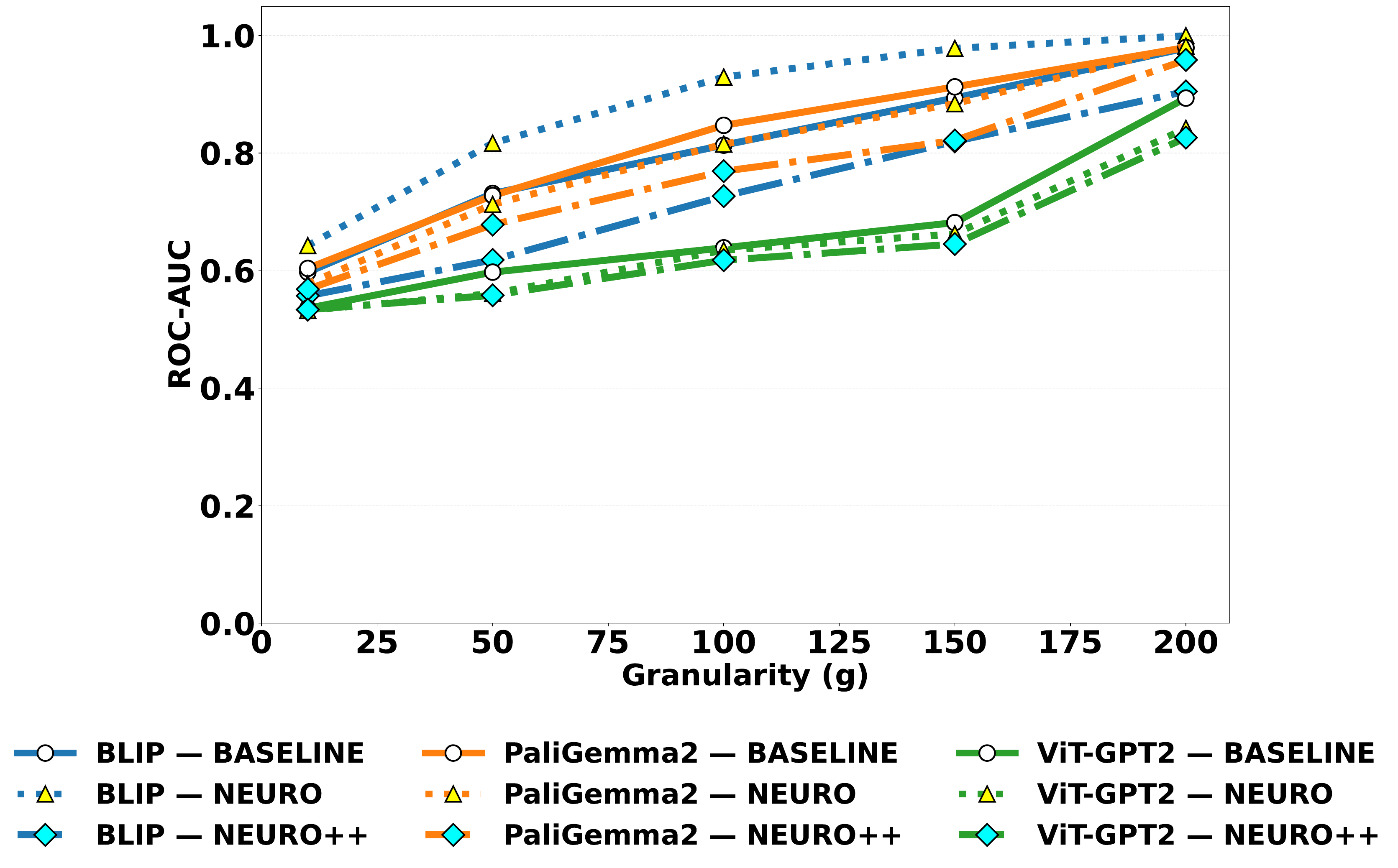}
         \caption{ROC-AUC with NoCaps Dataset}
         \label{fig:att_acc_nocaps}
     \end{subfigure}
        \caption{Plots showing attack success rates in ROC-AUC.}
        \label{fig:attack_accuracy}
\end{figure*}

\subsection{\textcolor{black}{Evaluation Metrics}}\label{subsec:eval_metrics}
In this section, we discuss the main evaluation metrics used in our experiments to measure the success of our attacks.

\textbf{ROUGE-2 Metric:}
We evaluate the lexical similarity of the generated captions using the \emph{ROUGE-2} metric~\cite{sharma2018conceptual, you2016image}. 
The general ROUGE framework is the most widely used evaluation metric for text summarization, serving as a content-based metric that measures the similarity between a generated text and a reference by quantifying overlapping n-grams \cite{barbella2022rouge, lin2004rouge}.

\textbf{MPNet Metric:}
We also evaluated the similarity between the generated and reference captions using \emph{MPNet} cosine similarity, an embedding-based semantic encoder. Semantic sentence embedding encodes sentences into vectors, such that when two sentences are close in this vector space, it means they have similar meanings~\cite{wieting2020bilingual, song2020mpnet, zeng2022contrastive, wajid2024deep}.

\textbf{ROC-AUC Metric:}
To provide a more precise assessment of the performance of our attack, we measure its success rate using the ROC-AUC area under the ROC Curve \cite{carlini2022membership, watson2021importance}. This method gives a stronger measure of how the attack performs at different levels of true positives \cite{jegorova2022survey}. It directly addresses the key goal of distinguishing between members and non-members \cite{song2017machine}.

\begin{table*}[t]
\vspace{4em}
\centering
\caption{
\textcolor{black}{Performance comparison among \textsc{Baseline} and $\tau$-regularized neuroscience-inspired models (\textsc{Neuro} with $\tau=2$ and \textsc{Neuro++} with $\tau=3$) on the CC3M dataset. The table reports member and non-member similarity scores for MPNet and ROUGE-2, along with the resulting ROC-AUC values (mean$\pm$std) across BLIP, PaliGemma 2, and ViT-GPT2. 
}}
\begin{tabular}{ |p{1.0cm}| p{1.8cm}| p{1.8cm}| p{1.8cm}| p{1.8cm}|  p{1.8cm}| p{1.8cm}| p{2.0cm}|  }
\hline
 Dataset  &Model  & Threat Model   & MPNet  $\downdownarrows$ & MPNet  $\downdownarrows$ &ROUGE-2 $\downdownarrows$ &ROUGE-2$\downdownarrows$& ROC-AUC$\upuparrows$\\
   &  &    & (Member)   & (Non-Member)  &(Member) &(Non-Member) & \\
\hline

CC3M &BLIP  &\textsc{Baseline}  &$0.560$ &$0.522$ & $0.152$ & $0.110$ &$84.23 \pm 15.45$\\ 
 & &\textbf{\textsc{Neuro}} & $\mathbf{0.561}$ & $\mathbf{0.519}$	&$\mathbf{0.158}$	 &$\mathbf{0.113}$ & $\mathbf{85.43 \pm 15.15}$ \\ 
\rowcolor{gray!50}
 & &  \textbf{\textsc{Neuro++}} & $\mathbf{0.475}$ & $\mathbf{0.450}$	&$\mathbf{0.093}$ &$\mathbf{0.075}$	 & $\mathbf{76.52 \pm 14.75}$ \\ 
\hline

\hline
\hline 
 &PaliGemma 2  &\textsc{Baseline}  &$0.418$ &$0.397$ &$0.056$ &$0.046$ &$71.69 \pm 11.40$\\ 
 & &\textbf{\textsc{Neuro}} & $\mathbf{0.486}$ & $\mathbf{0.462}$	&$\mathbf{0.079}$ &$\mathbf{0.068}$	 & $\mathbf{74.45 \pm 14.14}$ \\ 
\rowcolor{gray!50}
 & &  \textbf{\textsc{Neuro++}} & $\mathbf{0.451}$  & $\mathbf{0.436}$ 	&$\mathbf{0.073}$   &$\mathbf{0.063}$	 & $\mathbf{68.33 \pm 12.06}$ \\ 
\hline

\hline
\hline 
 &ViT-GPT2  &\textsc{Baseline}  &$0.363$ &$0.371$ &$0.041$ &$0.038$ &$48.13 \pm 16.11$\\ 
 & &\textbf{\textsc{Neuro}} & $\mathbf{0.363}$  & $\mathbf{0.372}$	&$\mathbf{0.042}$    &$\mathbf{0.038}$	 & $\mathbf{48.11 \pm 17.64}$ \\ 
\rowcolor{gray!50}
 & &  \textbf{\textsc{Neuro++}} & $\mathbf{0.363}$  & $\mathbf{0.371}$	&$\mathbf{0.043}$    &$\mathbf{0.037}$	 & $\mathbf{52.46 \pm 19.89}$ \\ 
\hline
\end{tabular}
\label{table:performance_cc3m}
\end{table*}

%% file: Sections/results.tex
\section{Experimental Results}
\label{sec:results}
In this section, we present our results after testing our three models, BLIP, PaliGemma 2, and ViT-GPT2 on datasets COCO, NoCaps, and CC3M. Our analysis covers different levels of topographic regularization, $\tau \in \{0, 2, 5\}$. We report these results using two methods: (i) similarity metrics (MPNet and ROUGE-2), and (ii) membership inference accuracy (ROC-AUC). Below, we discuss the patterns we observe.

\subsection{\textcolor{black}{Model Performance (Utility) Comparisons}}\label{subsec:perf_comp}
The three models exhibit distinct similarity profiles across the datasets. BLIP consistently achieves the highest MPNet and ROUGE-2 similarity scores at \(\tau=0\) (BASELINE), inferring much stronger alignment with reference captions. PaliGemma 2 has moderate scores, while ViT-GPT2 scores the lowest on most datasets. This ranking likely reflects their architectural design differences: BLIP's contrastive pretraining helps it achieve better lexical and semantic alignment \cite{li2022blip}, whereas ViT-GPT2 generates more varied, less specific captions \cite{ghosh2024exploring}. 
Additionally, higher similarity scores are associated with higher memorization. BLIP has the highest average similarity and the highest BASELINE ROC-AUC. To illustrate how generated captions differ between baseline models $\tau=0$ and neuro-inspired regularized models $\tau=3$ for member and non-member samples, we provide examples in Fig.~\ref{fig:coco_blip_image_cap_pairs}. Observe that the generated captions with \textsc{Neuro} VLMs are very similar to the reference caption, even under MIA attacks. This indicates the model utility is not significantly compromised in the \textsc{Neuro} VLMs.

In Table~\ref {table:performance_coco}, BLIP shows a (\textbf{94.00\%}) BASELINE ROC-AUC while ViT-GPT2 has the lowest at (\textbf{55.51\%}). This trend is also similar in Table~\ref {table:performance_cc3m}. In Table~\ref{table:performance_NoCaps}, although BLIP still shows the highest BASELINE ROC-AUC at (\textbf{81.74\%}), PaliGemma 2 follows closely with (\textbf{81.44\%}), while ViT-GPT2 still has the lowest at (\textbf{68.48\%}). It shows that models with stronger image-text alignment tend to leak membership information.

\subsection{\textcolor{black}{MIA Attack Success Rates}}\label{subsec:acc_comp}
As the value of $\tau$ increases, MIA success rates generally decrease for most (model, dataset) pairs. For the three VLMs tested, the lowest ROC-AUC for attacks is consistently observed at $\tau=3$ (NEURO++) across all \textbf{COCO}, \textbf{NoCaps}, and \textbf{CC3M} datasets.
\par
Topographic regularization enhances resilience against MIA attacks by reducing the model's sensitivity to patterns within member sets. However, the extent of this improvement depends on the dataset. For instance, COCO and CC3M exhibit substantial reductions in ROC-AUC, whereas NoCaps shows more minor changes due to its diverse captioning style and weaker direct alignment~\cite{agrawal2019nocaps}. We display the full ROC-AUC curves for COCO, CC3M, and NoCaps in Fig.~\ref{fig:attack_accuracy}.
\begin{figure*}
     \centering
     \begin{subfigure}[]{0.3\textwidth}
         \centering
         \includegraphics[width=\textwidth]{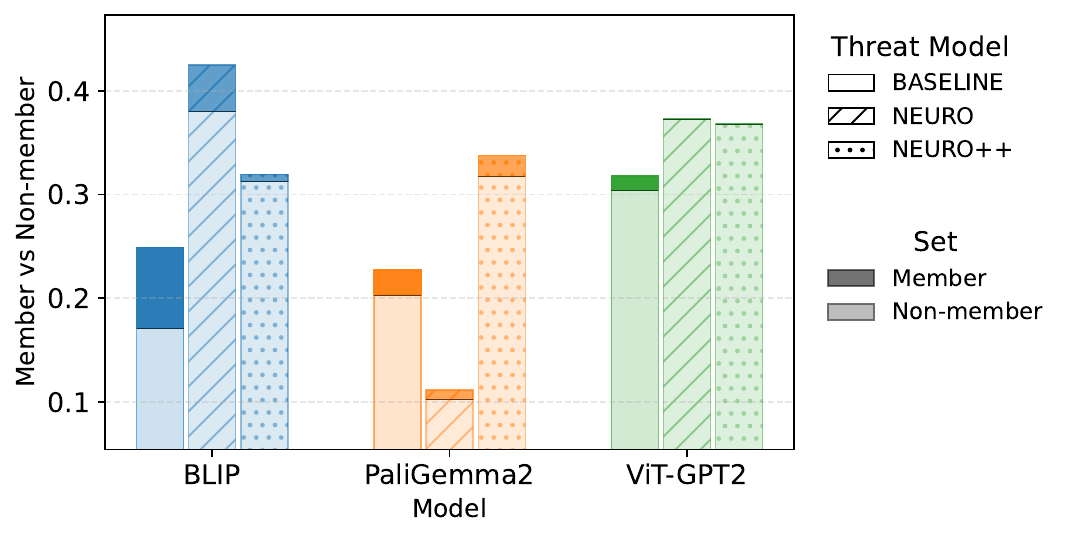}
         \caption{Similarity Means with COCO}
         \label{fig:sim_m_rouge_coco}
     \end{subfigure}
     \hfill
     \begin{subfigure}[]{0.3\textwidth}
         \centering
         \includegraphics[width=\textwidth]{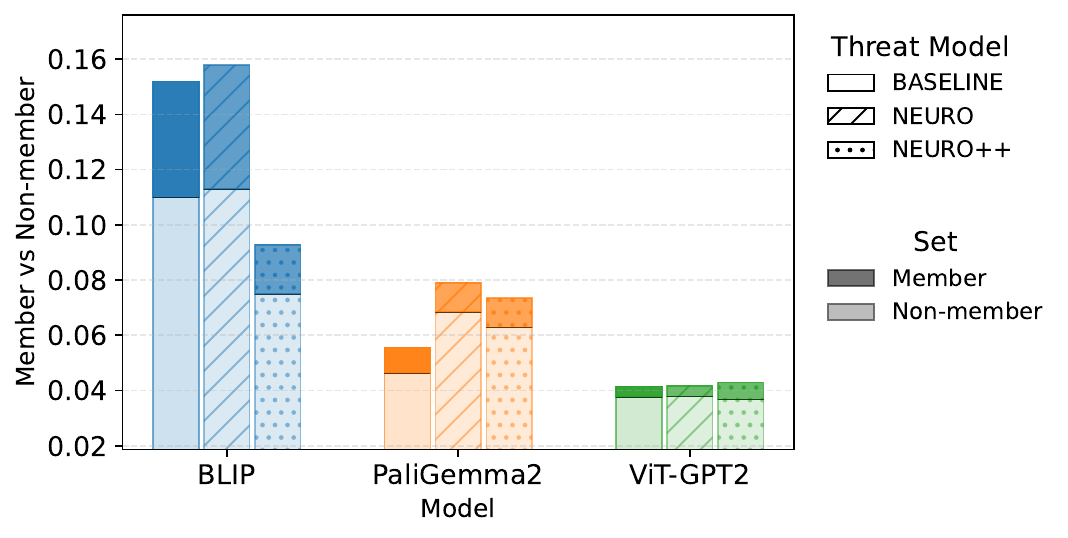}
         \caption{Similarity Means with CC3M}
         \label{fig:sim_m_rouge_cc3m}
     \end{subfigure}
     \hfill
     \begin{subfigure}[]{0.3\textwidth}
         \centering
         \includegraphics[width=\textwidth]{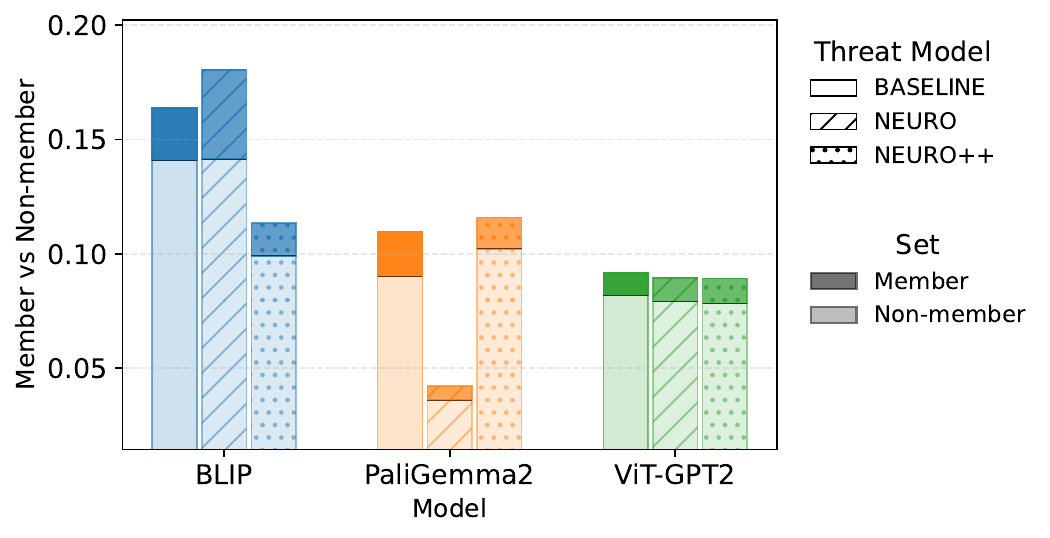}
         \caption{Similarity Means with NoCaps}
         \label{fig:sim_m_rouge_nocaps}
     \end{subfigure}
        \caption{Performance comparisons among \textsc{Baseline} and the $\tau$-regularized neuroscience-inspired models (\textsc{Neuro} with $\tau = 2$ and \textsc{Neuro++} with $\tau = 3$) in terms of similarity means (ROUGE-2) across multiple models (i.e., BLIP, PaliGemma 2, and ViT-GPT2) on three datasets\textemdash COCO, CC3M, and NoCaps.}
        \label{fig:sim_means_rouge}
\end{figure*}

\begin{figure*}
     \centering
     \begin{subfigure}[]{0.3\textwidth}
         \centering
         \includegraphics[width=\textwidth]{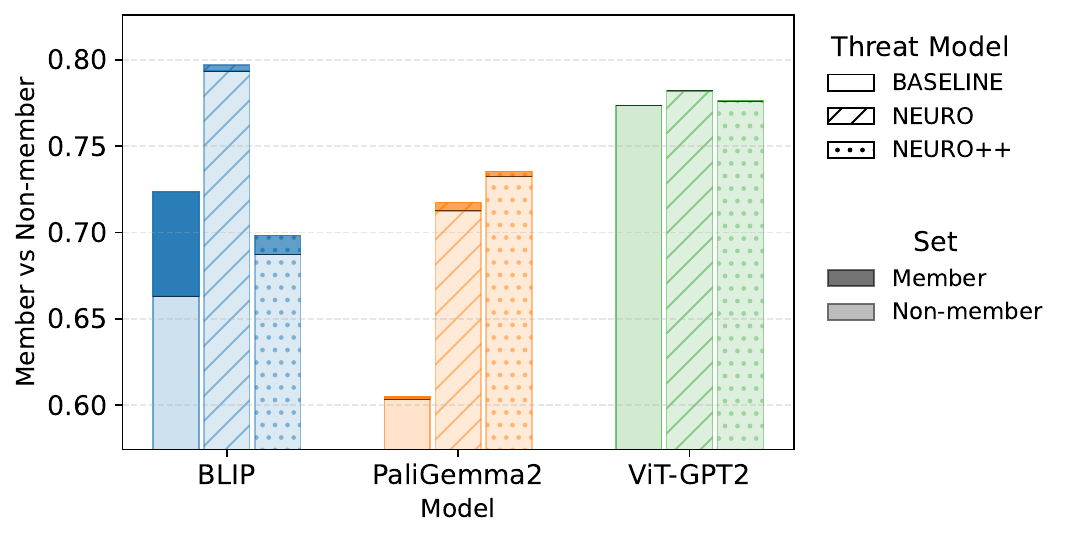}
         \caption{Similarity Means with COCO}
         \label{fig:sim_m_mpnet_coco}
     \end{subfigure}
     \hfill
     \begin{subfigure}[]{0.3\textwidth}
         \centering
         \includegraphics[width=\textwidth]{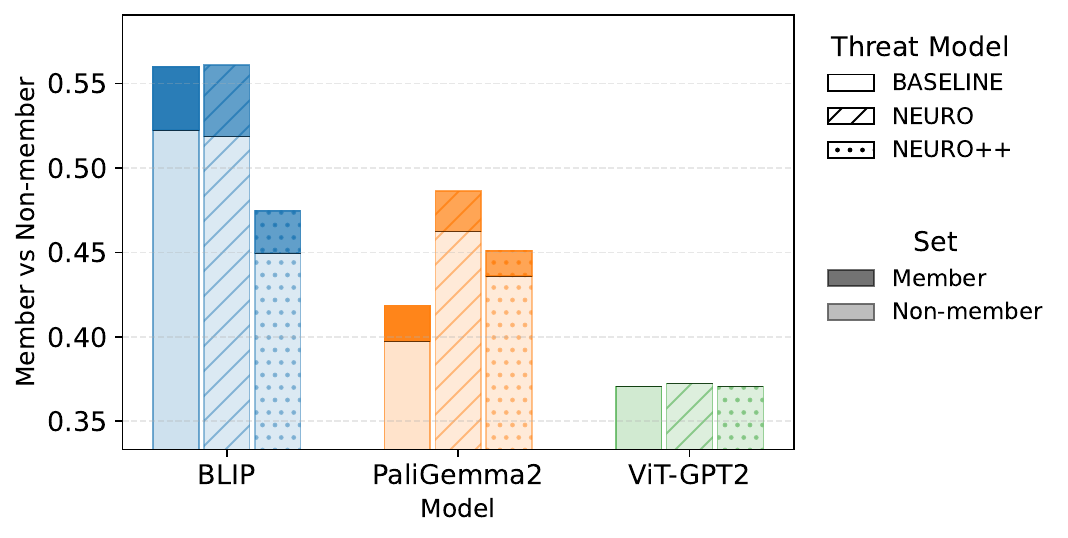}
         \caption{Similarity Means with CC3M}
         \label{fig:sim_m_mpnet_cc3m}
     \end{subfigure}
     \hfill
     \begin{subfigure}[]{0.3\textwidth}
         \centering
         \includegraphics[width=\textwidth]{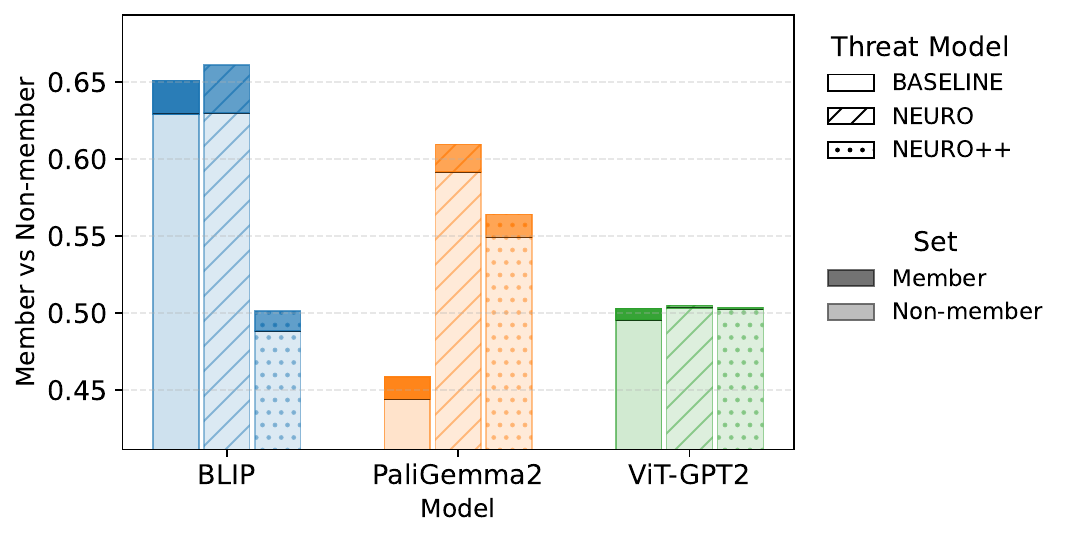}
         \caption{Similarity Means with NoCaps}
         \label{fig:sim_m_mpnet_nocaps}
     \end{subfigure}
        \caption{Performance comparisons among \textsc{Baseline} and the $\tau$-regularized neuroscience-inspired models (\textsc{Neuro} with $\tau = 2$ and \textsc{Neuro++} with $\tau = 3$) in terms of similarity means (MPNet) across multiple models (i.e., BLIP, PaliGemma 2, and ViT-GPT2) on three datasets\textemdash COCO, CC3M, and NoCaps.}
        \label{fig:sim_means_mpnet}
\end{figure*}

\subsection{\textcolor{black}{Impact of Neuro-Inspired Topological Regularization}}\label{subsec:topo_reg}
Fig.~\ref{fig:sim_means_rouge} summarizes how lexical similarity changes in terms of member and non-member sets across $\tau$. It shows the average ROUGE-2 scores for all models and datasets.
Increasing the value of $\tau$ affects both similarity metrics and the success of the attack. Fig.~\ref{fig:sim_means_mpnet} shows how the similarity means of MPNet also change in terms of member and non-member sets across $\tau$ for each dataset and model.  \par
In the COCO and CC3M datasets, MPNet and ROUGE-2 scores slightly change when $\tau$ increases from 0 to 3.
For example, in COCO for member sets, MPNet decreases from $0.723$ (BASELINE) to $0.698$ (NEURO++), and ROUGE-2 changes from $0.249$ to $0.319$. For non-member sets MPNet changes from $0.663$ (BASELINE) to $0.687$ (NEURO++), and ROUGE-2 changes from $0.171$ to $0.312$.
This indicates NEURO++ models can achieve similar or slightly better similarity scores between generated and reference captions, indicating no utility (model performance) degradation.

We can observe a similar trend in the NoCaps dataset with similar or slightly changed similarity scores (MPNet and ROUGE-2), which indicates NEURO++ does not compromise performance significantly.
We observe that the neuro-inspired models (BLIP) achieve lower ROC-AUC scores, indicating the resilience of the models against MIA attacks. For example, in the COCO dataset, NEURO and NEURO++ achieve ~24\% and ~30.5\% ROC-AUC drops, compared to the BASELINE. Similarly, across the CC3M and NoCaps datasets and other models (PaliGemma 2 and ViT-GPT2), we observe a drop in attack performance, i.e., ROC-AUC. This indicates higher resilience of the neuro-inspired models against MIAs.
\noindent\textbf{NEURO ($\tau{=}2$) vs.\ NEURO++ ($\tau{=}3$).}
A direct comparison between $\tau{=}2$ and $\tau{=}3$ shows a similar pattern based on the dataset. For COCO and CC3M, NEURO++ generally produces lower MPNet and ROUGE-2 scores than NEURO. PaliGemma 2 on COCO, however, shows an opposite trend, with NEURO++ producing higher similarity scores. This means there is no significant change in similarity; however, the ROC-AUC decreases in most scenarios, improving privacy beyond what NEURO provides.

On NoCaps, NEURO and NEURO++ usually produce very similar MPNet and ROUGE-2 scores, with only a slight drop in ROC-AUC. While NEURO++ offers attack resilience in datasets with structured captions, it has a limited impact on more diverse datasets.
Neuro-inspired regularization smooths internal representations and reduces overfitting to specific training examples. This mechanism narrows the similarity gap between members and non-members, thereby improving resiliency against attacks~\cite{deb2025toponets, pomi2018tensor, handjaras2015topographical, maxwell2020mapping}.

\subsection{\textcolor{black}{Ablations}}\label{subsec:ablation}
\begin{figure}
     \centering
     \begin{subfigure}[]{0.23\textwidth}
         \centering
         \includegraphics[width=\textwidth]{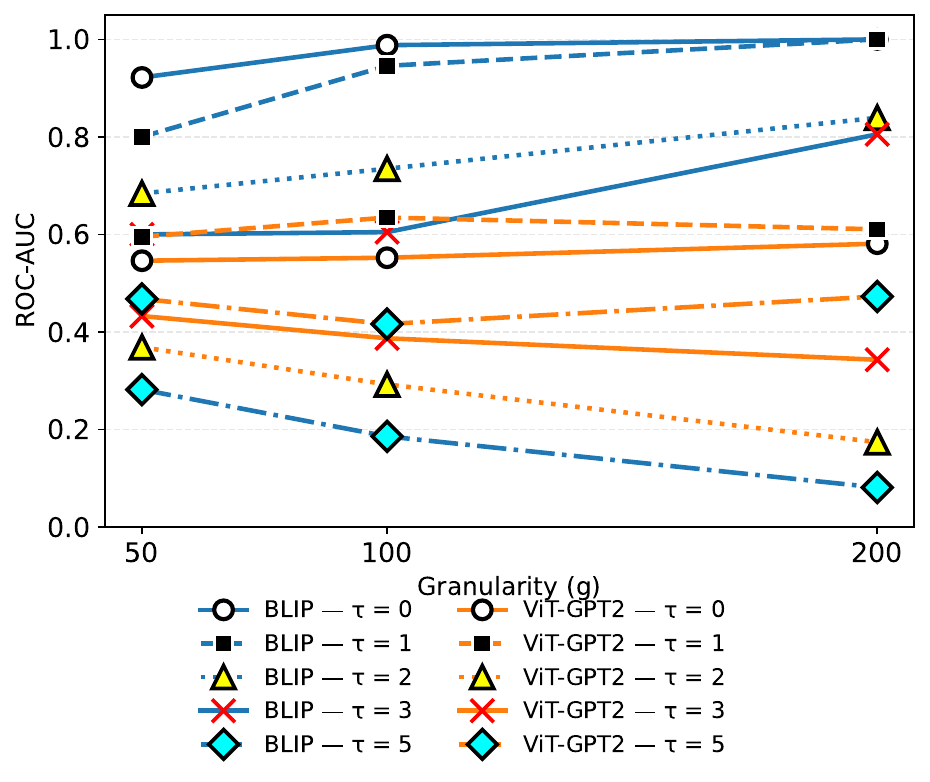}
         \caption{ROC-AUC on COCO}
         \label{fig:abla_att_acc_coco}
     \end{subfigure}
     \begin{subfigure}[]{0.23\textwidth}
         \centering
         \includegraphics[width=\textwidth]{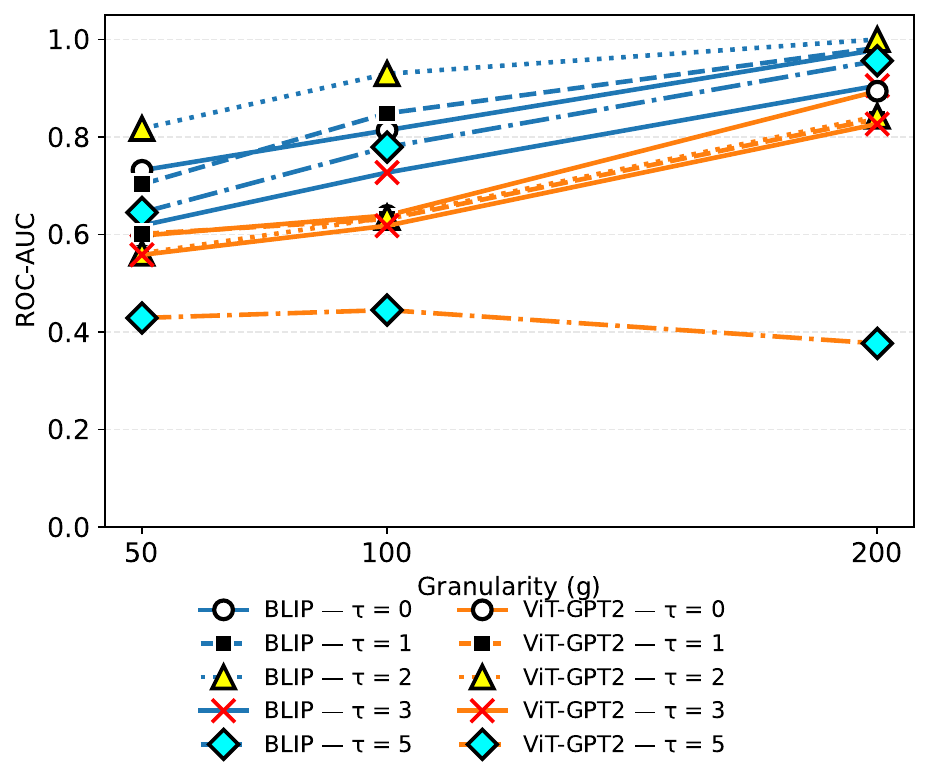}
         \caption{ROC-AUC on NoCaps}
         \label{fig:abla_att_acc_nocaps}
     \end{subfigure}
        \caption{Impact of $\tau$ on COCO and NoCaps with BLIP and ViT-GPT2 under
        $\tau \in \{0,1,2,3,5\}$ and granularities $g \in \{50,100,200\}$.
        }
        \label{fig:ablation_att_acc}
\end{figure}

To understand how topological regularization affects the attack setup, we tested it on two datasets (COCO and NoCaps) and two models (BLIP and ViT-GPT2). We varied the regularization strength $\tau$ with values \(\{0, 1, 2, 3, 5\}\) and the granularity \(g\) with options \(\{50, 100, 200\}\). For each combination of dataset, model, \(\tau\), and \(g\), we averaged the attack success rate with two similarity metrics (MPNet and ROUGE-2) and plotted the results in Fig.~\ref{fig:ablation_att_acc}.

\paragraph{Effect of granularity ($g$)}
The study shows that for both datasets, higher granularity leads to higher attack success, especially for BLIP. For example, on COCO with BLIP at $\tau=0$, the ROC-AUC increases from $g=50$ to nearly $100\%$ at $g=200$. When $\tau=1$, the attack success is slightly lower at $g=50$ and $g=100$, but it returns to $100\%$ at $g=200$. On the NoCaps dataset, BLIP and ViT-GPT2 show a similar pattern, where attacks strengthen as $g$ increases. However, ViT-GPT2 shows some unpredictable variations in its performance as $\tau$ changes. The study confirms that more finely grained attacks (larger $g$) are more successful at exploiting membership signals.

\paragraph{Effect of $\tau$ on COCO vs.\ NoCaps}
Increasing the value of $\tau$ from $0$ to $1$ reduces the ROC-AUC on BLIP in COCO at moderate granularities like $g=50$ and $g=100$. Higher values of $\tau$ ($\tau=2$, NEURO, and $\tau=3$ NEURO++) further lower the attack performance through different granularities. ViT-GPT2 shows noticeable changes between $\tau=0$ and $\tau=1$ on COCO. However, these differences are less noticeable than with BLIP, indicating a minor but consistent improvement in privacy. \par
On NoCaps, the effect of $\tau$ is weaker. BLIP shows an inconsistent trend, with $\tau=0$ having a weaker attack rate than higher $\tau$. The curves for $\tau=2$ and $\tau=3$ are identical throughout all granularities. It shows that NEURO and NEURO++ do not offer extra benefits over each other on BLIP. For ViT-GPT2, the attack rates are almost identical at lower $\tau$ for granularities $50$ and $100$, but a moderate gap in attack success reduction is observed when $g$ increases to $200$. The attack success is significantly reduced at $\tau=3$, and even more when $g$ increases from $100$ to $200$.

\begin{table*}[t]
\vspace{4em}
\centering
\caption{
\textcolor{black}{Performance comparison among \textsc{Baseline} and $\tau$-regularized neuroscience-inspired models (\textsc{Neuro} with $\tau=2$ and \textsc{Neuro++} with $\tau=3$) on the NoCaps dataset. The table reports member and non-member similarity scores for MPNet and ROUGE-2, along with the resulting ROC-AUC values (mean$\pm$std) across BLIP, PaliGemma 2, and ViT-GPT2. 
}}
\begin{tabular}{ |p{1.0cm}| p{1.8cm}| p{1.8cm}| p{1.8cm}| p{1.8cm}|  p{1.8cm}| p{1.8cm}| p{2.0cm}|  }
\hline
 Dataset  &Model  & Threat Model   & MPNet  $\downdownarrows$ & MPNet  $\downdownarrows$ &ROUGE-2 $\downdownarrows$ &ROUGE-2$\downdownarrows$& ROC-AUC$\upuparrows$\\
   &  &    & (Member)   & (Non-Member)  &(Member) &(Non-Member) & \\
\hline

NoCaps &BLIP  &\textsc{Baseline}  &$0.651$ &$0.629$ & $0.164$ & $0.141$ &$81.74 \pm 13.09$\\ 
 & &\textbf{\textsc{Neuro}} & $\mathbf{0.661}$ & $\mathbf{0.630}$	&$\mathbf{0.181}$	 &$\mathbf{0.141}$ & $\mathbf{87.35 \pm 13.91}$ \\ 
\rowcolor{gray!50}
 & &  \textbf{\textsc{Neuro++}} & $\mathbf{0.501}$ & $\mathbf{0.488}$	&$\mathbf{0.113}$ &$\mathbf{0.099}$	 & $\mathbf{72.55 \pm 13.99}$ \\ 
\hline

\hline
\hline 
 &PaliGemma 2  &\textsc{Baseline}  &$0.458$ &$0.444$ &$0.110$ &$0.090$ &$81.44 \pm 14.44$\\ 
 & &\textbf{\textsc{Neuro}} & $\mathbf{0.610}$ & $\mathbf{0.592}$	&$\mathbf{0.042}$ &$\mathbf{0.036}$	 & $\mathbf{79.45 \pm 14.87}$ \\ 
\rowcolor{gray!50}
 & &  \textbf{\textsc{Neuro++}} & $\mathbf{0.564}$  & $\mathbf{0.549}$ 	&$\mathbf{0.116}$   &$\mathbf{0.102}$	 & $\mathbf{75.93 \pm 14.06}$ \\ 
\hline

\hline
\hline 
 &ViT-GPT2  &\textsc{Baseline}  &$0.503$ &$0.495$ &$0.092$ &$0.082$ &$68.48 \pm 13.85$\\ 
 & &\textbf{\textsc{Neuro}} & $\mathbf{0.505}$  & $\mathbf{0.503}$	&$\mathbf{0.090}$    &$\mathbf{0.079}$	 & $\mathbf{64.66 \pm 14.97}$ \\ 
\rowcolor{gray!50}
 & &  \textbf{\textsc{Neuro++}} & $\mathbf{0.503}$  & $\mathbf{0.502}$	&$\mathbf{0.089}$    &$\mathbf{0.078}$	 & $\mathbf{63.63 \pm 15.93}$ \\  
\hline
\end{tabular}
\label{table:performance_NoCaps}
\end{table*}

\ignore{====================
\begin{figure*}[t]
    \centering
    \includegraphics[width=0.7
    \textwidth]{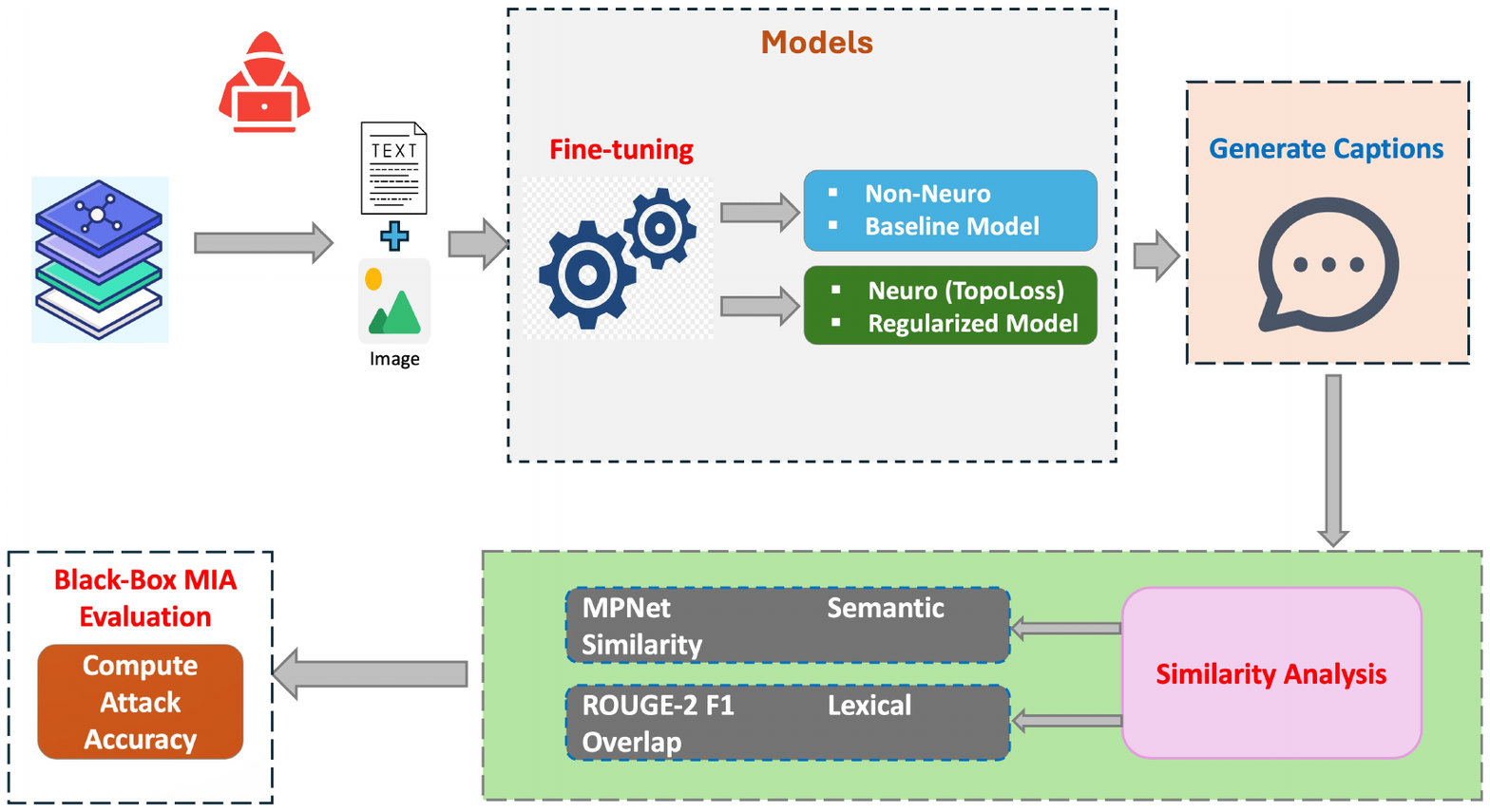}
    \caption{\textcolor{red}{A detailed taxonomy of Adversarial and Privacy Vulnerability on Diffusion Models.}}
    \label{fig:vlm_overview}
\end{figure*}
==========================}


%% file: Sections/conclusion.tex
\section{Conclusion}
\label{sec:conclusion}
In this paper, we demonstrated that biologically inspired topographic regularization is an effective method for mitigating MIAs in VLMs. Our studies show that $\tau$ regularization reduces the gap in similarity between member and non-member captions, thereby improving resilience against MIAs. Note that \textsc{Neuro} VLMs maintain model utility (similar similarity scores as baselines), while improving resilience against MIAs. These findings suggest several promising directions, including identifying optimal $\tau$ for privacy-utility trade-offs, exploring \textit{white-box} MIAs on VLMs. These steps are essential for creating deployable, privacy-preserving \textsc{Neuro} VLMs in real-life agentic AI applications.